\documentclass[fleqn,10pt]{wlscirep}
\usepackage[utf8]{inputenc}
\usepackage[T1]{fontenc}
\usepackage{lineno}
\usepackage{subcaption}

\title{
TACK Tunnel Data (TTD): A Benchmark Dataset for Deep Learning-Based Defect Detection in Tunnels
}

\author[1,*]{Andreas Sjölander}
\author[2]{Valeria Belloni}
\author[1]{Robel Fekadu}
\author[1]{Andrea Nascetti}
\affil[1]{Civil and Architectural Engineering, KTH Royal Institute of Technology, Stockholm, Sweden}
\affil[2]{Department of Civil, Building and Environmental Engineering, Sapienza University of Rome, Rome, Italy}

\affil[*]{corresponding author: Andreas Sjölander(asjola@kth.se)}

\begin{abstract}
Tunnels are essential elements of transportation infrastructure, but are increasingly affected by ageing and deterioration mechanisms such as cracking. Regular inspections are required to ensure their safety, yet traditional manual procedures are time-consuming, subjective, and costly. Recent advances in mobile mapping systems and Deep Learning (DL) enable automated visual inspections. However, their effectiveness is limited by the scarcity of tunnel datasets. This paper introduces a new publicly available dataset containing annotated images of three different tunnel linings, capturing typical defects: cracks, leaching, and water infiltration. The dataset is designed to support supervised, semi-supervised, and unsupervised DL methods for defect detection and segmentation. Its diversity in texture and construction techniques also enables investigation of model generalization and transferability across tunnel types. By addressing the critical lack of domain-specific data, this dataset contributes to advancing automated tunnel inspection and promoting safer, more efficient infrastructure maintenance strategies. 
\end{abstract}
\begin{document}

\flushbottom
\maketitle

\thispagestyle{empty}

\section*{Background \& Summary}

Tunnels are critical components of infrastructure systems and are typically designed for a technical lifespan of 100 years or more. However, in many countries, existing tunnels are ageing, as most construction materials degrade over time due to natural and mechanical deterioration mechanisms such as cracking and water infiltration.
To maintain structural integrity throughout their intended lifespan, regular inspection and maintenance are essential. Consequently, infrastructure owners must continuously plan and optimize monitoring and maintenance strategies. The aim is to minimize the risk of failure that can result in catastrophic consequences, including loss of life, while also limiting downtime that disrupts transportation networks and causes significant economic losses. 
Traditionally, inspections are performed on-site by experts with basic tools such as hammers and headlamps. For detailed inspections of the concrete lining, skylifts are often required to access the tunnel surface closely. This approach is time-consuming, labor-intensive, prone to human errors, and inherently subjective \cite{Sjolander2023}. Today, sensors can be easily placed on Mobile Mapping Systems (MMS), enabling automatic collection of infrastructure data. Specifically, high-resolution cameras mounted on an MMS can be easily used to scan the tunnel and collect a large number of images that depict the tunnel lining. This allows the tunnel to be remotely inspected in the office, which reduces the tunnel closing time and improves inspection protocols \cite{Sjolander2023, Belloni2020, Foria_2022}. However, the inspection is still performed mainly visually.
Therefore, the key challenge is to achieve sufficient accuracy in damage detection, mainly cracks, using images collected with high-resolution cameras and automatic image processing techniques.   

Automatic damage detection methods can be categorized into two main approaches: traditional image processing \cite{Mohan_2018} and Machine Learning/Deep Learning (ML/DL) techniques \cite{Belloni2023}. 
Among these, DL methods have recently gained increasing attention due to their superior capabilities in representation learning. DL techniques are generally categorized into three groups: supervised, semi-supervised, and unsupervised learning.
Supervised learning typically relies on large annotated datasets to train models such as Convolutional Neural Networks (CNNs) and Transformers for classification, object detection, or segmentation tasks. Supervised models require extensive and labour-intensive labelling for training and testing the model effectively, since the quality and quantity of labeled data highly affect the performance of supervised learning. 
Semi-supervised learning seeks to overcome these limitations by combining a small set of labelled data with a larger volume of unlabeled data. Techniques such as consistency regularization, pseudo-labeling, and self-training help propagate label information from annotated samples to unlabeled ones. Finally, unsupervised methods such as autoencoders, clustering algorithms, diffusion models, and generative adversarial networks aim to eliminate the need for labeled data, aiming to enhance generalization across diverse datasets.  
Supervised learning has been extensively explored for crack detection, primarily using CNNs \cite{Liu2019, Liu2019DeepCrack, Zou_2019DeepCrack, Liu2020, Raza2022, Lan_2022, Liu2022, Belloni2023, Ouyang2023}, Transformers \cite{Liu_2021_ICCV, Asadi_Shamsabadi_2022, Chu_2024}, or hybrid architectures that combine both \cite{Xiang_2023}. In contrast, semi-supervised \cite{Wang_2021, Jian_2024} and unsupervised \cite{unsupervised, Ma_2024} methods have only recently begun to receive attention. 

Due to the recent spread of supervised DL architectures and their need for labelled data, many datasets are now publicly available for image classification, object detection, and semantic segmentation. Image classification is useful only to determine whether a defect is present (damage/no damage), but it does not provide specific information on the damage. Object detection aims at detecting the location of defects using bounding boxes. Finally, semantic segmentation identifies defects at the pixel level, allowing their precise delineation and measurement, including their length, width, and area \cite{Belloni2020}. The majority of these datasets focus only on crack detection in pavements, bridges and buildings. Other datasets investigate multi-class damage detection mainly in pavements and bridges 
(Table \ref{tab:SOTA_datasets}). 
Finally, a few works \cite{OmniCrack30k_github, OmniCrack30k, crackseg9k_github, crackseg9k} merge some of the data described in Table \ref{tab:SOTA_datasets} to provide larger datasets.
\begin{table}[h!]
\centering
\begin{tabular}{lccccccc}

\textbf{Dataset} & \textbf{Task} & \textbf{Images} & \textbf{Resolution} 
& \textbf{Infrastructure} & \textbf{Number of classes} \\
\hline
Concrete crack \cite{dataset1, dataset1a} & Classification & 40000 & 227 $\times$ 227 & Building & 1 - Crack\\
Crack-detection \cite{dataset2, dataset2b} & Classification & 6069 & 224 $\times$ 224 & Bridge & 1 - Crack \\
SDNET2018 \cite{Maguire, DORAFSHAN2018} & Classification & 56000  & 256 $\times$ 256 & Bridge/Building/Pavement & 1 - Crack\\
SUT-Crack \cite{SUT-Crack, Sabouri2023} & Classification & 25563 & 200 $\times$ 200 & Pavement & 1 - Crack\\
SUT-Crack \cite{SUT-Crack, Sabouri2023} & Segmentation & 130 & 3024 $\times$ 4032 & Pavement & 1 - Crack\\
SUT-Crack \cite{SUT-Crack, Sabouri2023} & Object detection & 130 & 3024 $\times$ 4032 & Pavement & 1 - Crack\\
BCL \cite{BCL, Ye2021} & Segmentation & 11000 & 256 $\times$ 256 & Bridge & 1 - Crack\\
DeepCrack \cite{DeepCrack_github, dataset1b} & Segmentation & 537 & Mixed & Mixed & 1 - Crack\\
CFD \cite{CFD_github, shi2016automatic, cui2015pavement} & Segmentation & 118  & 480 $\times$ 320 & Pavement & 1 - Crack\\
CrackNJ156 \cite{CrackNJI56, Xu2022} & Segmentation & 156  & $1734 \times 1734$  & Pavement & 1 - Crack\\
Masonry \cite{Masonry, Dais2021} & Segmentation & 11491 & 224 $\times$ 224 & Masonry structure & 1 - Crack \\
NHA12D \cite{NHA12D, NHA12D2} & Segmentation & 80  & 1920 $\times$ 1080  & Pavement & 1 - Crack \\
GAPs 10m \cite{GAPs, stricker2021road} & Segmentation & 20 & 5030 $\times$ 11505 & Pavement & 22 \\
GAPs v2 \cite{GAPs, stricker2019imp} & Object detection & 2468 & 1920 $\times$ 1080 & Pavement & 5  \\
S2DS \cite{benz2022image, S2DS} & Segmentation & 743 & 1024 $\times$ 1024 & Concrete Surface & 6  
\\
CODEBRIM \cite{CODEBRIM, CODEBRIM_github, Mundt_2019_CVPR} & Object detection & 1590 & Mixed & Bridge & 5 \\
dacl10k \cite{dacl10k_dataset, dacl10k} & Segmentation & 9920 & Mixed & Bridge & 19 \\
\hline
\end{tabular}
\caption{\label{tab:SOTA_datasets}Publicly available datasets for single and multi-class damage detection. 
}

\end{table}



In general, most available datasets focus only on defects in asphalt pavements, concrete buildings and bridges, which makes the prediction more challenging when focusing on a different domain, that is, tunnels.    
Tunnels present a unique environment in terms of light conditions, where the false detection of cables, installations, joints or other objects can be very common. Moreover, depending on the technique adopted to build the tunnel, the characteristics of the tunnels can be completely different. As an example, rock tunnels are usually supported with a combination of rock bolts and a concrete lining. In hard rock conditions, thin linings of fibre-reinforced shotcrete (sprayed concrete) is typically used, while a cast concrete lining with ordinary steel bar reinforcement is used in areas with poor rock conditions \cite{Sjölander2020}. This leads to different textures in the collected images, which has a high impact on damage detection tasks.   

Although data collection of cracks or defects in structures is time-consuming, roads and bridges are accessible in most cases, which facilitates data collection. For tunnels, data collection is typically an expensive operation, and extensive planning with owners is required to gain access to the tunnel. Due to these limitations, only a few studies focused on damage detection in tunnels. Some studies \cite{tunnel1, tunnel2, tunnel3, tunnel4, Ouyang2023} used tunnel data for training and testing, but they are not publicly available. Xue et al., 2020 \cite{water_tunnel1, water_tunnel2} presented a dataset on water leakage in a shield tunnel. Finally, Feng et al., 2024 \cite{FENG2024, FENG2024_2} provided a dataset for object detection and image classification of leakage, spalling, and cracks in Shanghai metro lines. In general, the lack of tunnel datasets makes it extremely difficult to train and test supervised, semi-supervised and unsupervised networks accurately. 

To address this gap, this paper presents a unique publicly available dataset consisting of labelled images from three different tunnels containing cracks, leaching, and water, which are typical in hard rock tunnels. Among these, the presence of a crack does not always directly compromise tunnel safety, but it often initiates and accelerates concrete deterioration mechanisms, notably the corrosion of steel fibres \cite{Galan2019}. Therefore, reliable crack detection is crucial for accurately assessing and predicting structural degradation.  
For supervised, semi-supervised and unsupervised networks, this dataset enables training and testing of various architectures. Moreover, the limitation of tunnel datasets provides another challenge in terms of damage detection since current DL models present limitations related to generalization. This is the result from these models being trained and tested mainly on small and very specific datasets, which greatly limits the model's ability to handle domain shifts, such as varying materials and context. For both supervised, semi-supervised and unsupervised models, this dataset offers the opportunity to assess generalization and transferability issues across different tunnel types. 

\section*{Methods}

Below, the data collection and the labelling work are presented in more detail.

\subsection*{Data Collection}
Data were collected within the Tunnel Automatic CracK Detection (TACK) project, a research initiative aimed at developing an autonomous inspection method to improve tunnel monitoring and assessment through the integration of deep learning and photogrammetry \cite{Belloni2020, Belloni2023, Belloni2025, Sjolander2023b, Belloni_thesis}.  
The inspection procedure proposed in the project is based on an MMS equipped with LiDAR (Light Detection And Ranging) sensors, panoramic cameras and a rig of high-resolution cameras, which collect data from the tunnel environment. LiDAR data are used to generate a digital twin of the infrastructure, while images are annotated to create datasets of tunnel defects to train and test DL models. Subsequently, a novel photogrammetric algorithm \cite{Belloni2023, Sjolander2023b, Belloni2025b} is used to track the evolution of cracks over time using the collected images.  Finally, a structural condition assessment is conducted to identify areas that require visual inspection. 

\begin{figure}[h!]
	\centering
	\subfloat
	{\includegraphics[width=70mm]{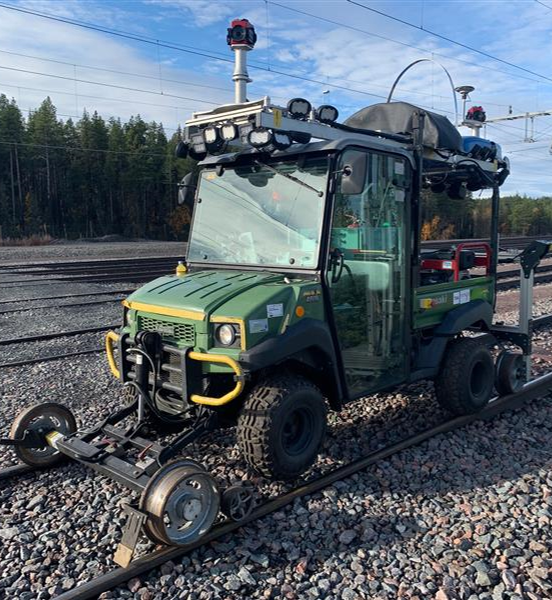}}
    \hspace{0.2mm}
	\subfloat
	{\includegraphics[width=70mm]{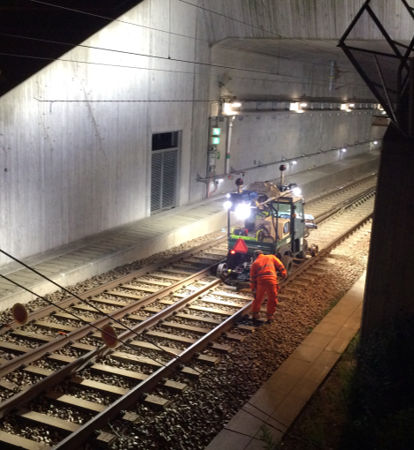}}
	\caption{Mobile mapping system developed by WSP Sweden \cite{WSP2024}. From \cite{Sjolander2023, Belloni2025}.}
	\label{fig:mobile_mapping}
\end{figure}

The data presented in this paper were captured using an MMS developed by WSP Sweden \cite{WSP2024} at a speed of approximately 5 km/h. The system consists of six LiDAR sensors, two panoramic cameras and a rig with seven InfraRed (IR) cameras combined with IR flashes \cite{Belloni2020} (Figure \ref{fig:mobile_mapping}).
The dataset focuses on the high-resolution images collected using CMOS (Sony IMX264) sensors mounted on the MMS with a resolution of 2448 $\times$ 2048 pixels, a pixel size of 3.45 microns, a focal length of 12 mm and a global shutter function that ensures high-quality imaging. 

\begin{figure}[h!]
	\centering
    	\subfloat[Tunnel A]
	{\includegraphics[width=55mm]{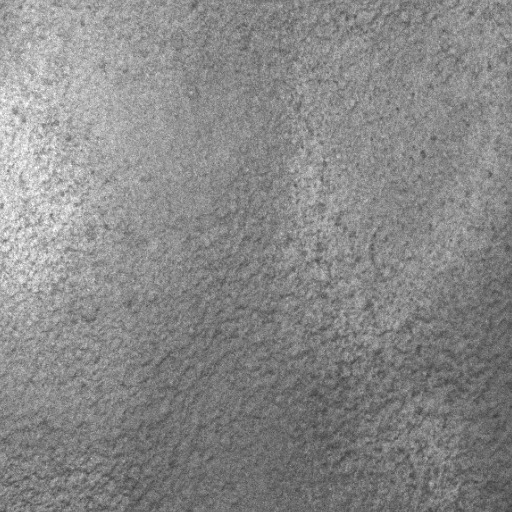}\label{fig:HR-imagesA}}
    \hfill
    	\subfloat[Tunnel B]
	{\includegraphics[width=55mm]{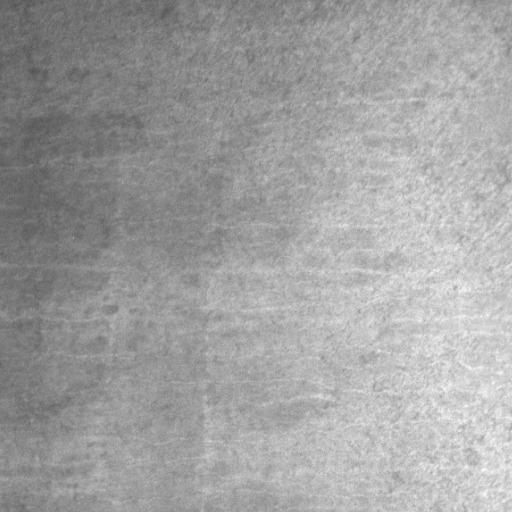}\label{fig:HR-imagesB}}
    \hfill
        	\subfloat[Tunnel C]
    {\includegraphics[width=55mm]{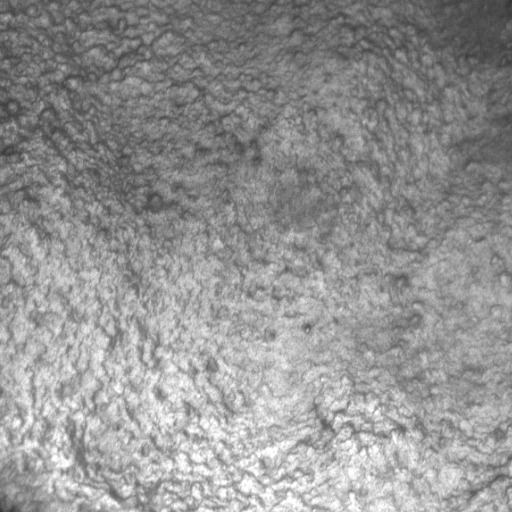}\label{fig:HR-imagesC}}
	\caption{Example of high-resolution images depicting the surface of the three tunnels.}
	\label{fig:HR-images}
\end{figure}

The images are collected inside three different tunnels. Tunnel A (TA) represents a modern constructed highway tunnel in hard rock. Here, fibre-reinforced shotcrete (sprayed concrete) is first sprayed directly on the rock surface, which constitutes the main rock support together with rock bolts. Thereafter, an inner shotcrete lining has been sprayed on top of a membrane attached to the rock. This creates a free-standing inner lining of shotcrete, and the purpose of this lining is to create a dry tunnel environment. Since the inner lining is sprayed against a membrane, the surface is smoother compared to shotcrete sprayed directly on rock.  Still, the texture is more coarse, which is a result of the application process in which shotcrete is sprayed with high pneumatic pressure and the use of set accelerators \cite{Sjölander2020} (Figure \ref{fig:HR-imagesA}). Tunnel B (TB) is constructed using pre-cast concrete elements, which result in a smooth surface similar to any other type of concrete structure (Figure \ref{fig:HR-imagesB}). Tunnel C (TC) is a rock tunnel supported by fibre-reinforced shotcrete sprayed directly on the rock surface. The tunnel was excavated using the drill and blast method, which resulted in a highly irregular rock surface and has, consequently, an irregular shotcrete surface (Figure \ref{fig:HR-imagesC}). 

\subsection*{Data Labelling}

Data labelling represents a crucial activity for the application of DL methods, since it is necessary to train and test DL architectures. Traditionally, data labelling has been manual, subjective, and very time-consuming. In addition, the quality of the annotations has a high impact on the performance of training and testing the networks. The introduction of annotation platforms has improved annotation tasks. These platforms are time-efficient since they typically offer a semi-autonomous labelling system based on a pre-trained segmentation model. An example of pre-trained models is the Segment Anything Model (SAM) developed by Kirillov et al. \cite{Kirillov2023}, which could be used to automatically segment cracks. However, preliminary studies indicate that while the model has good accuracy in labelling larger and clearly visible cracks, it may need to be fine-tuned to reach an acceptable accuracy in labelling narrow cracks that are less clearly visible \cite{Sjolander2025}.

\begin{figure}[ht]
\centering
\includegraphics[width=\linewidth]{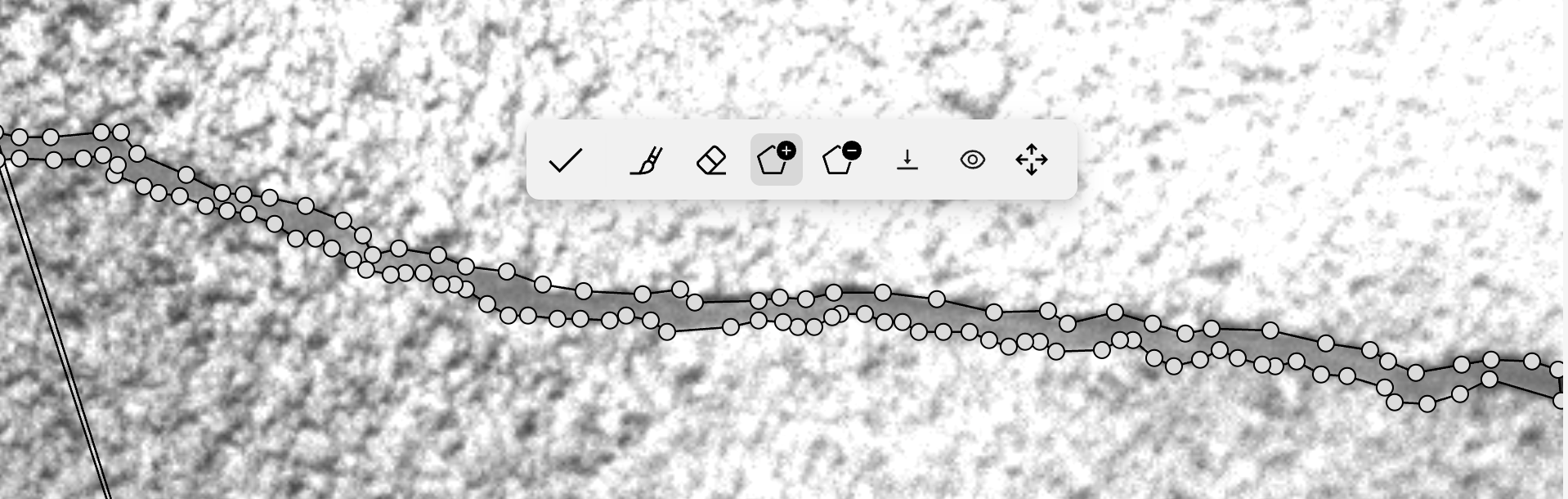}
\caption{Manual labelling of a crack in which each point represents a manually added point.}
\label{fig:labelling}
\end{figure}

In our work, we used SuperAnnotate to speed up image annotation. SuperAnnotate is a cloud platform developed to annotate, train and automate computer vision projects \cite{SuperAnnotate2024}. Thanks to various artificial intelligence features, the platform provides a powerful tool to increase efficiency with pre-trained segmentation models. Labelling features with large and distinct differences, such as a wet section of the concrete, is therefore very efficient and semi-autonomous, while labelling cracks is still mainly done manually.  An example of a manually segmented crack is shown in Figure \ref{fig:labelling}. Here, each dot around the exterior of the crack represents a manually added point, which clearly highlights the time-consuming process of image labelling.

As the first step in creating this dataset, all images were manually inspected to detect damage on the concrete surface and the original dataset was split into two groups: damage/no damage. Then, pre-labelling of the damaged imagery was performed. Here, damaged sections were outlined by one annotator and then evaluated and potentially re-assessed by a second annotator. Finally, pre-labelled imagery was annotated at the pixel level by one annotator and checked by a second one. In this way, we reduced errors related to subjective annotation, especially for tunnels supported with sprayed concrete excavated by drill and blast, where the texture of the concrete is rough and the surface is highly irregular due to variations in concrete thickness and the rock surface. This results in shadows, making the detection of cracks difficult.

During labelling, the priority was to identify cracks, as this is a common initiation of the degradation of concrete and fibres. Sections that showed clear signs of being damped were also marked, as this indicates that water is present on the rock surface and is penetrating the shotcrete through cracks or the porous system. Water often initiates or accelerates the deterioration of concrete and steel, and it should be monitored for this purpose. Finally, leaching was also labelled as a defect. This phenomenon is common in structures with one-sided water pressure (e.g., tunnels \cite{Sjölander2019, Galan2019}) and is characterised by a distinctive white precipitation on the concrete surface. If cracks are not present, leaching is normally a slow process that takes many years to initiate. However, depending on the extent of leaching in relation to when the tunnel was constructed, leaching could be an indicator of the presence of cracks in the concrete. 
In detail, the dataset contains the following categories:

\begin{itemize}
    \item Crack - wide and narrow cracks as well as cracks that were mainly identified through water, i.e. a damp section of the concrete that follows a typical irregular and narrow shape of a crack;
    \item Water - a damp section of the concrete;
    \item Leaching - a section on the surface that contains white precipitation.
\end{itemize}

\subsection*{Dataset description}
The dataset presented in this paper includes 785 images with cracks, 197 images with water and 316 images with leaching. The total number of images and distribution between each category are presented in Table \ref{tab:data}. The distribution of pixels between the categories is presented in Table \ref{tab:data_pixels}. 

\begin{table}[h!]
\centering
\begin{tabular}{lccccc}

\textbf{Dataset} & \textbf{Images}& \textbf{No damage} & \textbf{Crack}  & \textbf{Water} & \textbf{Leaching} \\
\hline
Tunnel A & 932 & 565 & 337 & 2 & 29\\
Tunnel B & 1536 & 1176 & 263 & 33 & 116\\
Tunnel C & 1306 & 848 & 185  & 162 & 171\\
Total           & 3774 & 2589 & 785 & 197 & 316\\
\hline
\end{tabular}
\caption{\label{tab:data}Description of the complete dataset, divided into three tunnel types and three classes.}
\end{table}
\begin{table}[h!]
\centering
\begin{tabular}{lcccccc}

\textbf{Dataset} & \textbf{Total Pixels} & \textbf{Crack}  & \textbf{Water} & \textbf{Leaching} \\
\hline
Tunnel A & 100 & 0.346  &0.009 &0.041\\
Tunnel B & 100 & 0.095  & 0.318 & 0.257\\
Tunnel C & 100 & 0.097  & 1.841 & 0.350\\
\hline
\end{tabular}
\caption{\label{tab:data_pixels}Percentage of pixels for each class for the three tunnel types and three classes.}
\end{table}

\begin{figure}[h!]
	\centering
	{\includegraphics[width=180mm]{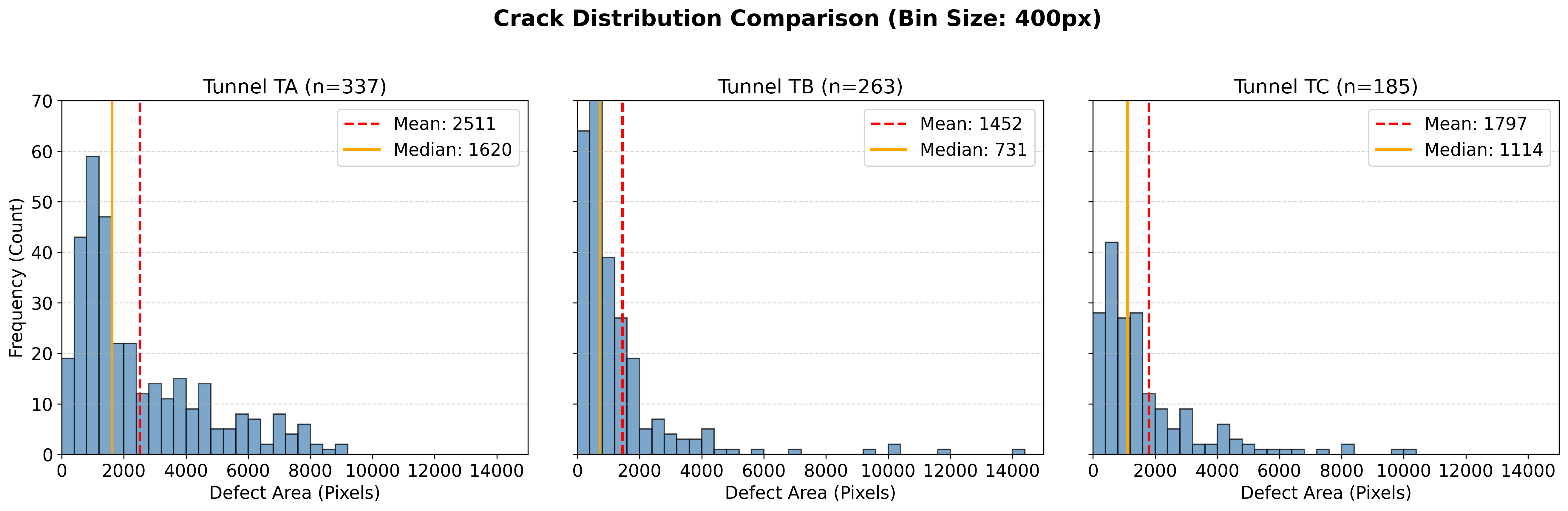}\label{fig:histogram_a}}
    \hspace{0.2mm}
	\caption{Crack size distribution in the three tunnels.}
	\label{fig:histogram}
\end{figure}

\begin{figure}[h!]
    \centering
    
    \begin{subfigure}[b]{0.45\textwidth}
        \centering
        \includegraphics[width=75mm]{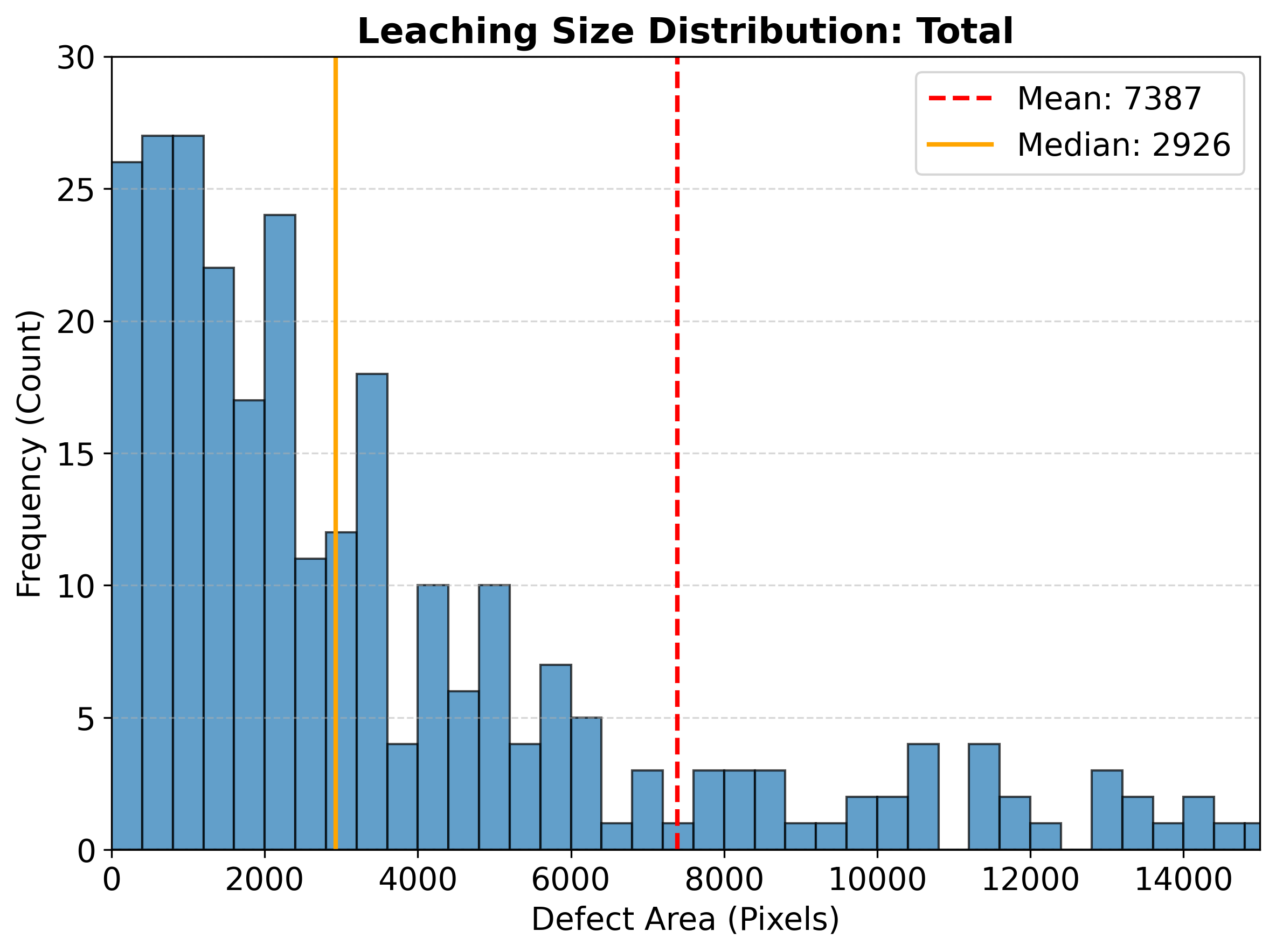}
    \end{subfigure}
    \hfill
    \begin{subfigure}[b]{0.45\textwidth}
        \centering
        \includegraphics[width=75mm]{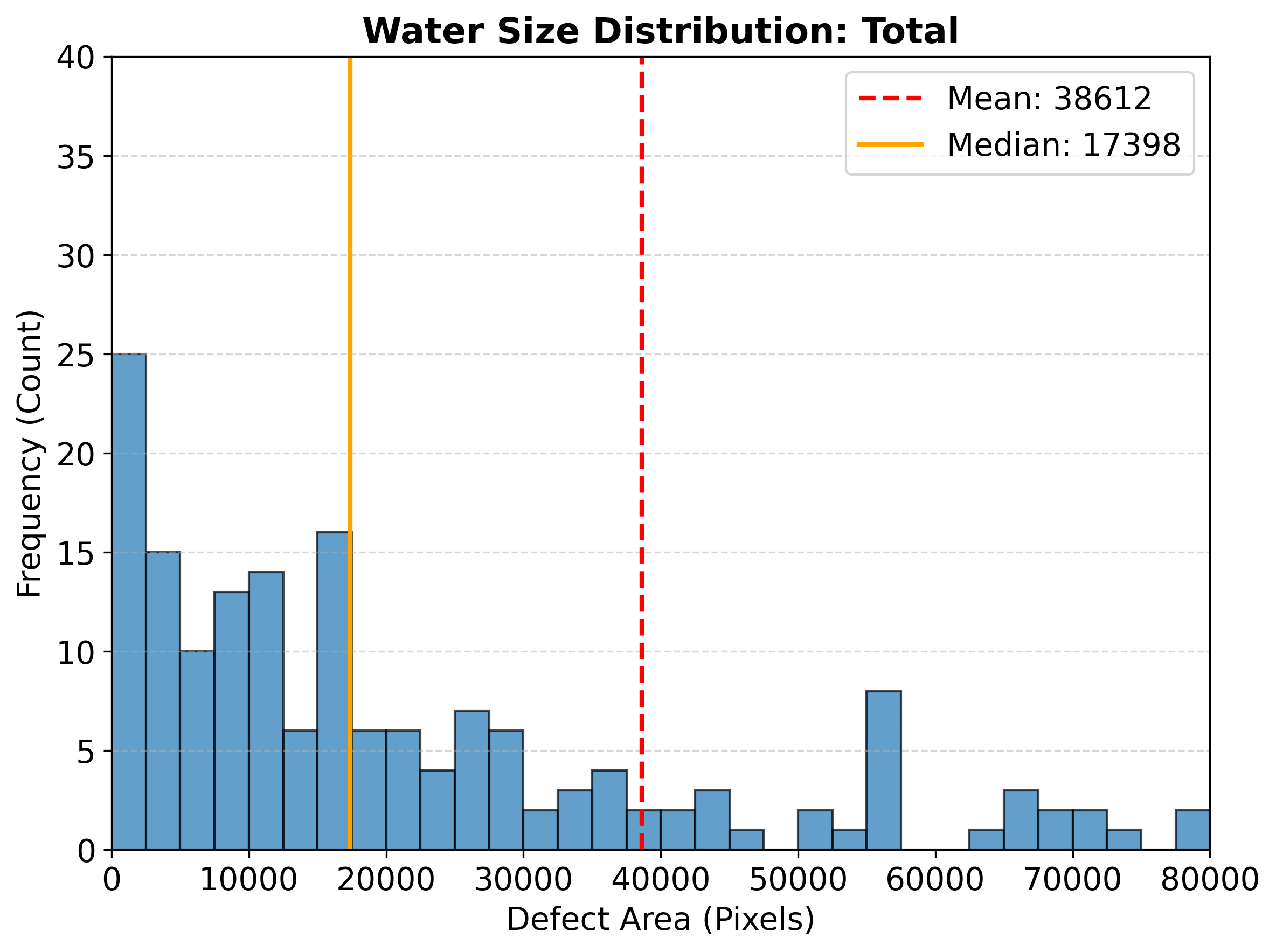}

    \end{subfigure}
    
    \vspace{1em} 
  
    \caption{Distribution of water and leaching in the three tunnels. }
    \label{fig:water_leaching}
\end{figure}

The main focus of this dataset is to provide labelled images of cracks in tunnels. Therefore, Figure \ref{fig:histogram} shows the histograms of the crack area in pixels for each tunnel. The figure shows that most cracks are small, which is also confirmed by the comparison between the median and mean value of the area. Moreover, TA has a higher frequency with larger cracks compared to TB and TC. The distribution of water and leaching is summarized for all tunnels in Figure \ref{fig:water_leaching}. Compared to cracks, these defects have a significantly larger area.

\clearpage
The dataset is publicly available on Hugging Face \cite{TTD2025} and is composed of four folders:

\begin{itemize}
    \item \textbf{1\_python}: this folder contains Python scripts that calculate statistics for the dataset, create a classification dataset and train and evaluates a CNN model. All scripts and their task is summarized in "0 read me.py";
    \item \textbf{2\_model\_input}: contains .csv files that link images and masks used in the balanced dataset for model training. For each tunnel, four csv files exist. The files TA-Train, TA-Val and TA-Test were used to train and test the CNN model and contains a balanced number of cracked and uncracked images. The file TA-Dataset-Labels can be used to create a balanced classification dataset;
    \item \textbf{3\_img}: contains all images in grayscale. For training purposes, originally collected images were tiled in a size of 512 $\times$ 512 pixels. The original names of the images were retained, and indices A-D were assigned to the tiled images;
    \item \textbf{3\_mask}: contains the 512 $\times$ 512 pixel mask corresponding to each image in the 3\_img folder. Each defect was given a unique pixel value: crack = 40, water = 160 and leaching = 200. 
\end{itemize}

\begin{figure}[h!]
    \centering
    
    \begin{subfigure}[b]{0.3\textwidth}
        \centering
        \includegraphics[width=55mm]{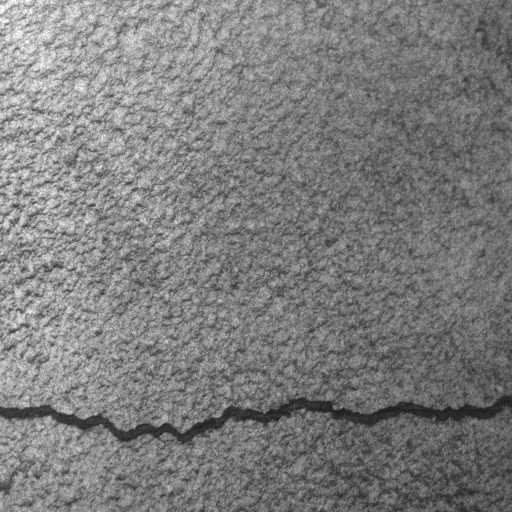}

    \end{subfigure}
    \hfill
    \begin{subfigure}[b]{0.3\textwidth}
        \centering
        \includegraphics[width=55mm]{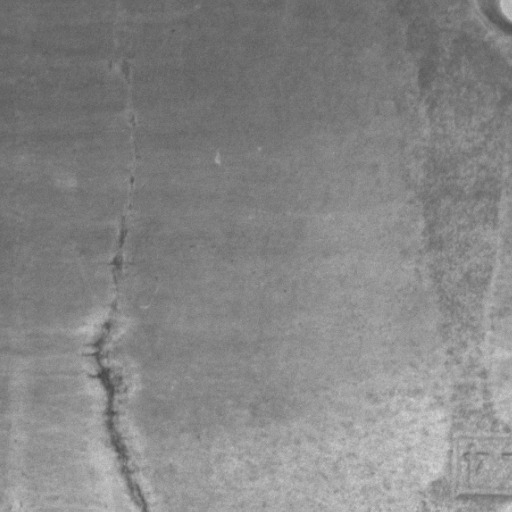}
        
    \end{subfigure}
    \hfill
    \begin{subfigure}[b]{0.3\textwidth}
        \centering
        \includegraphics[width=55mm]{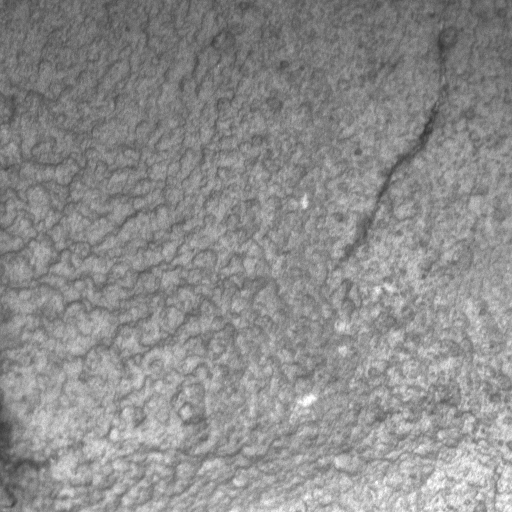}
    \end{subfigure}
    
    \vspace{0.2em} 
    
    \begin{subfigure}[b]{0.3\textwidth}
        \centering
        \includegraphics[width=55mm]{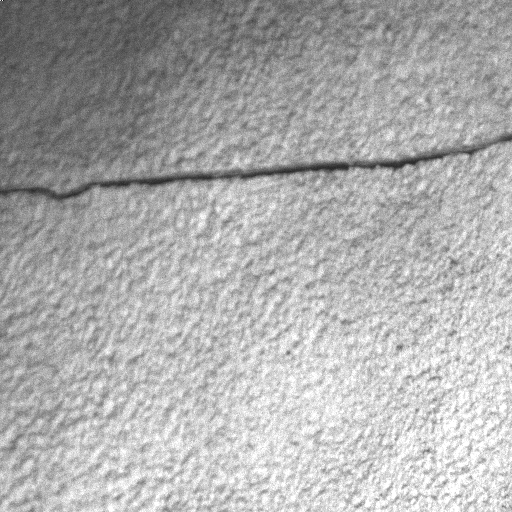}
    \end{subfigure}
    \hfill
    \begin{subfigure}[b]{0.3\textwidth}
        \centering
        \includegraphics[width=55mm]{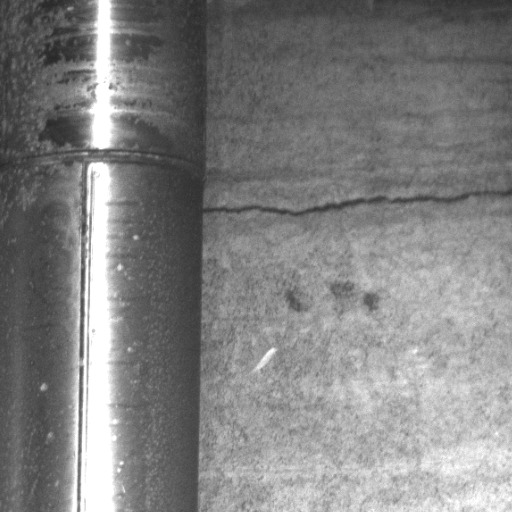}
    \end{subfigure}
    \hfill
    \begin{subfigure}[b]{0.3\textwidth}
        \centering
        \includegraphics[width=55mm]{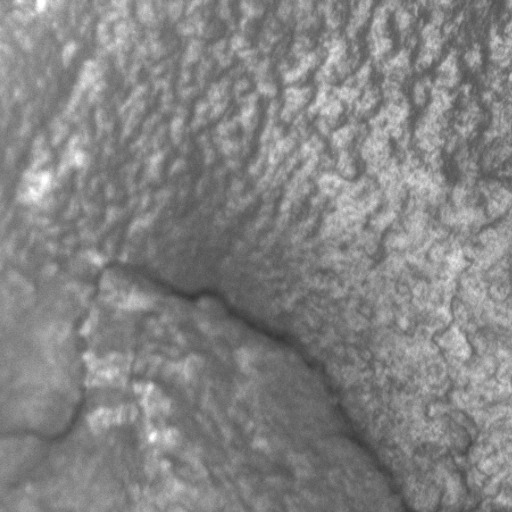}
    \end{subfigure}
    
    \vspace{0.2em}
    
    \begin{subfigure}[b]{0.3\textwidth}
        \centering
        \includegraphics[width=55mm]{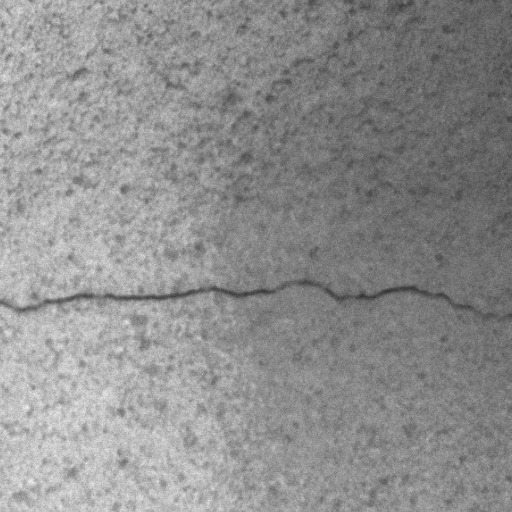}
        \caption{Tunnel A}
    \end{subfigure}
    \hfill
    \begin{subfigure}[b]{0.3\textwidth}
        \centering
        \includegraphics[width=55mm]{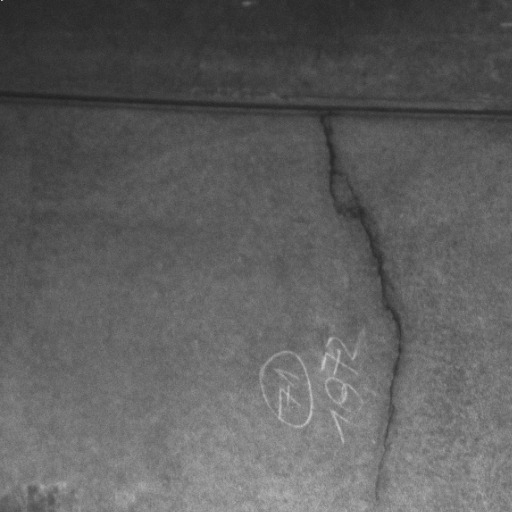}
        \caption{Tunnel B}
    \end{subfigure}
    \hfill
    \begin{subfigure}[b]{0.3\textwidth}
        \centering
        \includegraphics[width=55mm]{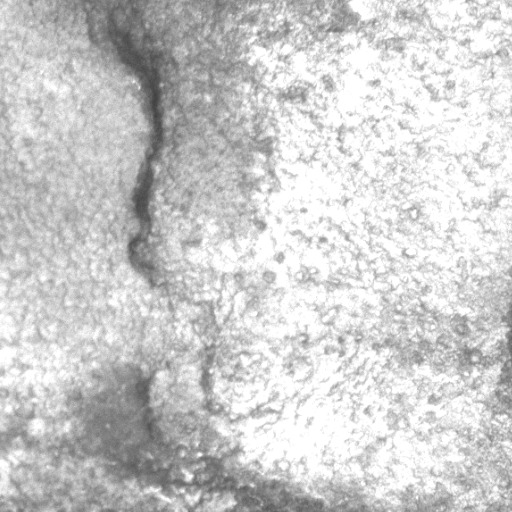}
        \caption{Tunnel C}
    \end{subfigure}
    
    \caption{Examples of easily detected cracks from Tunnel A (a), Tunnel B (b) and Tunnel C (c). }
    \label{fig:damage_1}
\end{figure}

\begin{figure}[t]
    \centering
    
    \begin{subfigure}[b]{0.3\textwidth}
        \centering
        \includegraphics[width=55mm]{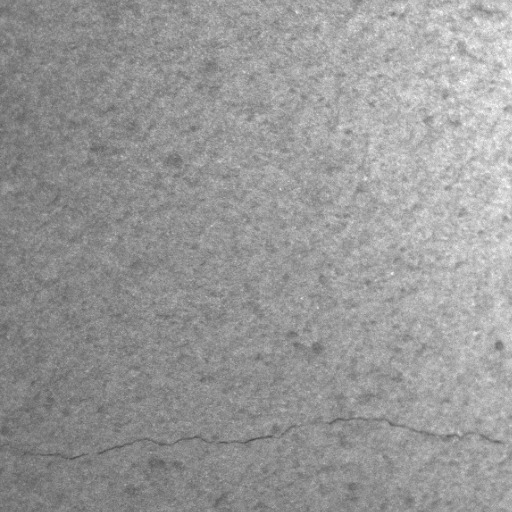}
    \end{subfigure}
    \hfill
    \begin{subfigure}[b]{0.3\textwidth}
        \centering
        \includegraphics[width=55mm]{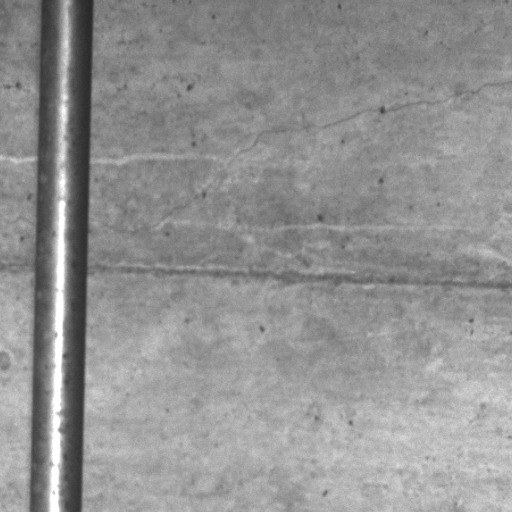}
    \end{subfigure}
    \hfill
    \begin{subfigure}[b]{0.3\textwidth}
        \centering
        \includegraphics[width=55mm]{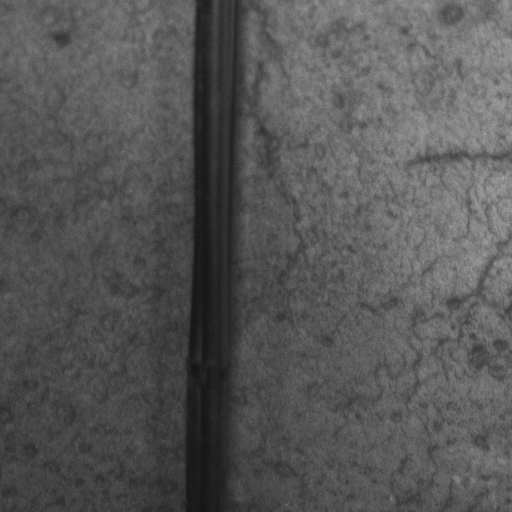}
    \end{subfigure}
    
    \vspace{0.2em} 
    
    \begin{subfigure}[b]{0.3\textwidth}
        \centering
        \includegraphics[width=55mm]{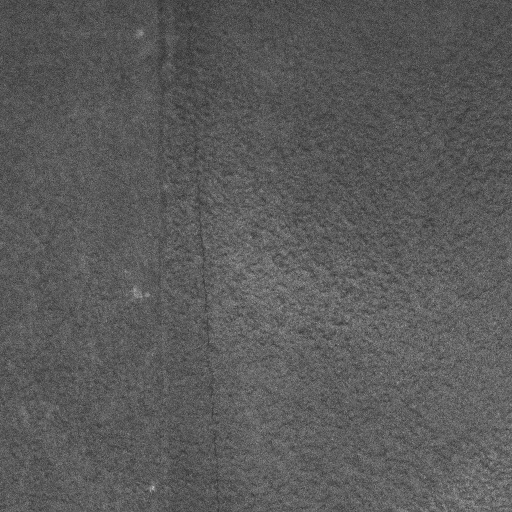}
    \end{subfigure}
    \hfill
    \begin{subfigure}[b]{0.3\textwidth}
        \centering
        \includegraphics[width=55mm]{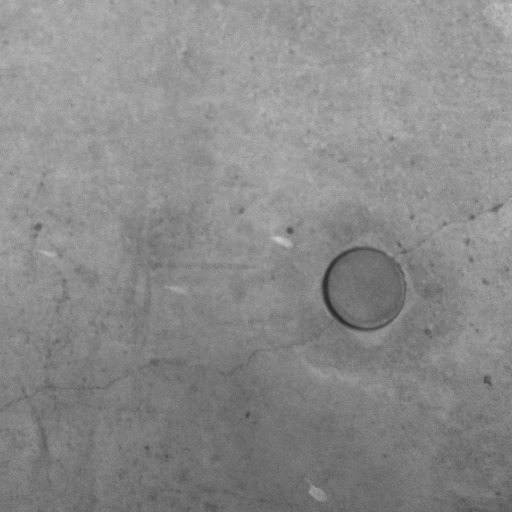}
    \end{subfigure}
    \hfill
    \begin{subfigure}[b]{0.3\textwidth}
        \centering
        \includegraphics[width=55mm]{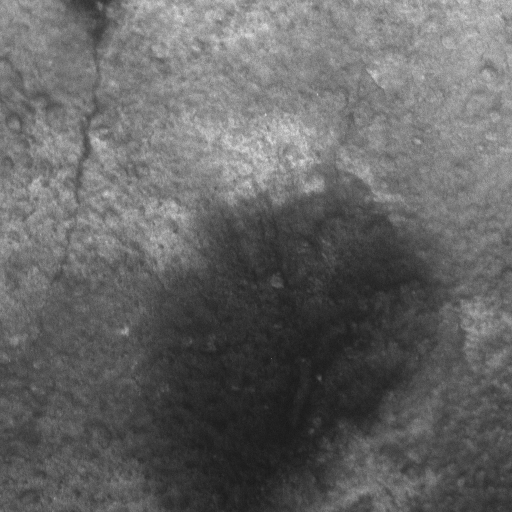}
    \end{subfigure}
    
    \vspace{0.2em}
    
    \begin{subfigure}[b]{0.3\textwidth}
        \centering
        \includegraphics[width=55mm]{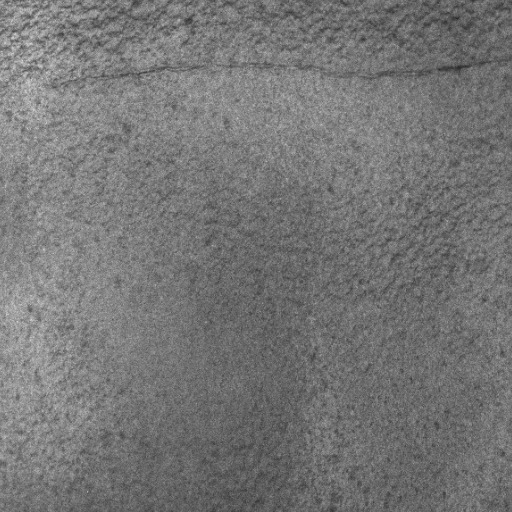}
        \caption{Tunnel A}
    \end{subfigure}
    \hfill
    \begin{subfigure}[b]{0.3\textwidth}
        \centering
        \includegraphics[width=55mm]{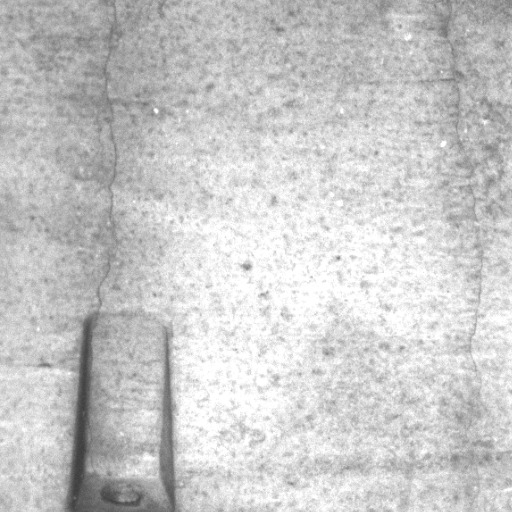}
        \caption{Tunnel B}
    \end{subfigure}
    \hfill
    \begin{subfigure}[b]{0.3\textwidth}
        \centering
        \includegraphics[width=55mm]{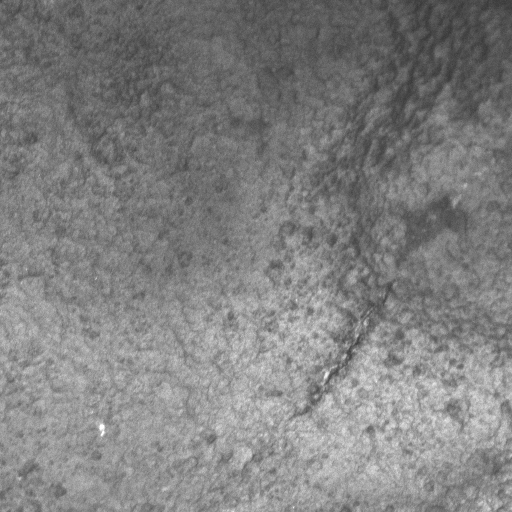}
        \caption{Tunnel C}
    \end{subfigure}
    
    \caption{Examples of more complex cracks to detect from Tunnel A (a), Tunnel B (b) and Tunnel C (c).}
    \label{fig:damage_2}
\end{figure}

\begin{figure}[t]
    \centering
    \begin{subfigure}[b]{0.3\textwidth}
        \centering
        \includegraphics[width=55mm]{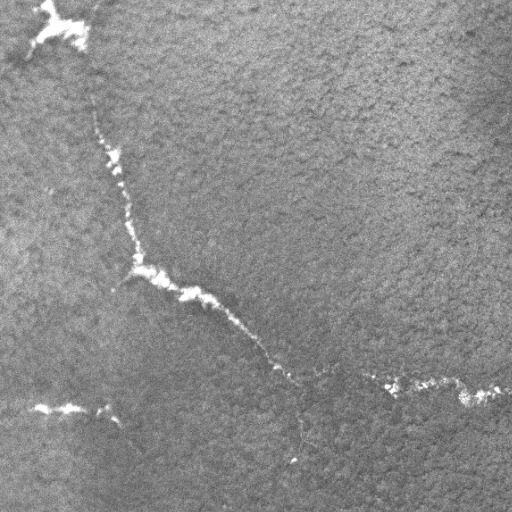}
    \end{subfigure}
    \hfill
    \begin{subfigure}[b]{0.3\textwidth}
        \centering
        \includegraphics[width=55mm]{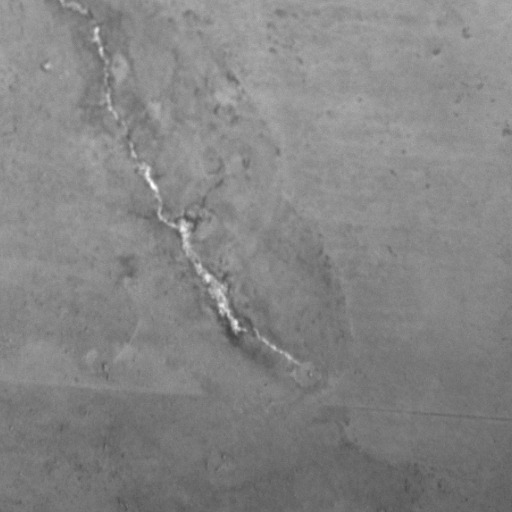}
    \end{subfigure}
    \hfill
    \begin{subfigure}[b]{0.3\textwidth}
        \centering
        \includegraphics[width=55mm]{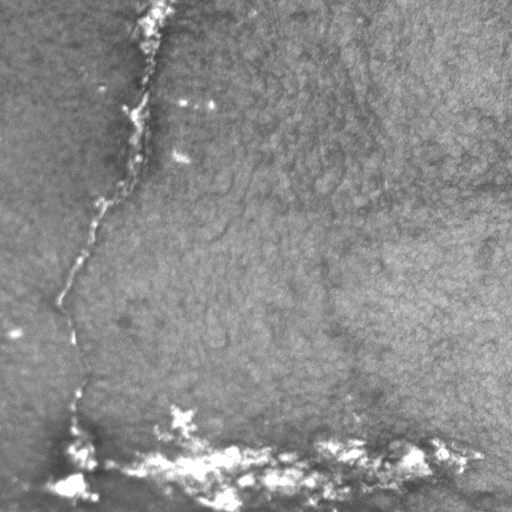}
    \end{subfigure}
    
    \vspace{0.2em}
    
    \begin{subfigure}[b]{0.3\textwidth}
        \centering
        \includegraphics[width=55mm]{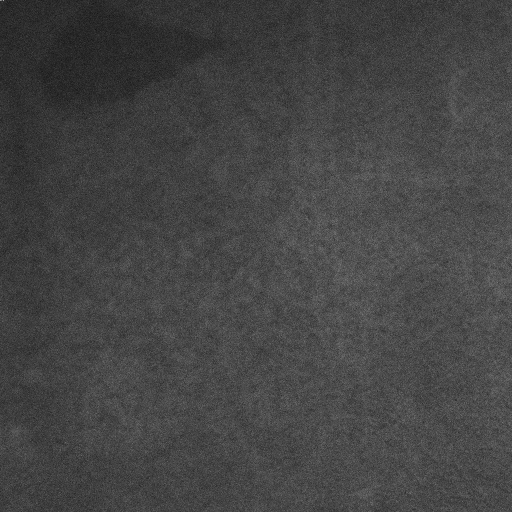}
        \caption{Tunnel A}
    \end{subfigure}
    \hfill
    \begin{subfigure}[b]{0.3\textwidth}
        \centering
        \includegraphics[width=55mm]{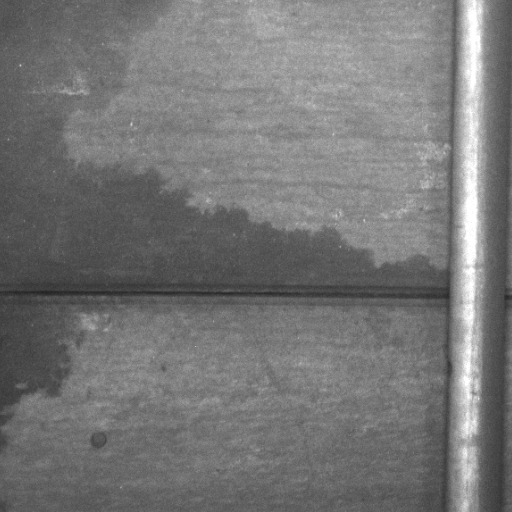}
        \caption{Tunnel B}
    \end{subfigure}
    \hfill
    \begin{subfigure}[b]{0.3\textwidth}
        \centering
        \includegraphics[width=55mm]{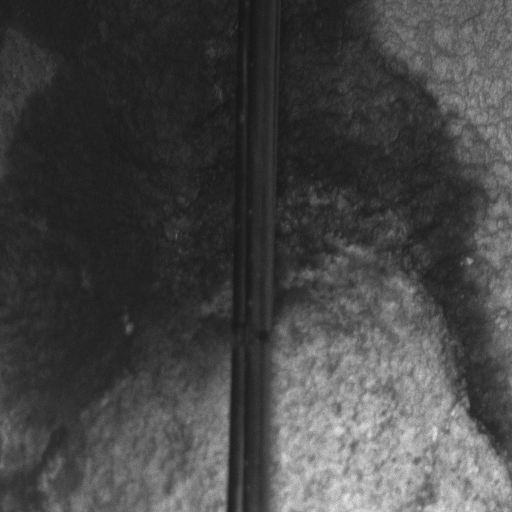}
        \caption{Tunnel C}
    \end{subfigure}
    
    \caption{Examples of images with leaching (top row) and water (bottom row) from Tunnel A (a), Tunnel B (b) and Tunnel C (c).}
    \label{fig:water_leacing_ex}
\end{figure}

The Python scripts use relative links to ensure mobility of the dataset. For all tunnels, a csv file named \textit{TX-Dataset-Label} was created. This file contains an equal number of image names with and without cracks. This is indicated with the labels and target \textit{crack / 1} and \textit{no crack / 0} in the file. These files are used to create the classification dataset and was used to split each tunnel dataset into training, testing and validation. These csv files are also available in \textbf{2\_model\_input} and were used for the validation of the dataset to ensure that the same images were used in each phase of training, validation and testing. An example of cracks from each tunnel is shown in Figure \ref{fig:damage_1} while Figure \ref{fig:damage_2} shows examples of cracks and other objects in the tunnel that could lead to false detection of cracks. Finally, Figure \ref{fig:water_leacing_ex} show examples of leaching and water from each tunnel.

\clearpage
\clearpage
\clearpage

\section*{Technical Validation}

Finding cracks and other defects in the concrete lining of a tunnel can be a subjective and complex process. In this work, all the labelled damages were visually verified by two of the authors to minimize subjective judgment. Both the high-resolution imagery and the overview imagery taken by the MMS panoramic cameras were used to verify the existence of the crack. Moreover, all images labelled as \textit{background} (for segmentation models) and \textit{no-crack} (for classification models) have been manually checked by two persons to verify that these images do not contain any defects (cracks, water or leaching for segmentation) or cracks (for classification). In case of disagreement, the final decision was made by the senior expert. In the case of cracks, uncertain cracks were more often kept than disregarded. The reason for this is that, for inspection purposes, it is better to evaluate and disregard potential cracks than completely miss them. However, assessing cracks is still a subjective task and it is recommended to review images with cracks using the classification script.

\subsection*{Validation of dataset}
To validate the quality of the labels in the dataset, a CNN model was trained and tested. Here, the focus was on crack detection since the majority of labelled images contain cracks. A single-class detection model was created using the \textit{fastai} \cite{fastai} Python package, which is built on top of \textit{torch} \cite{torch}. The U-Net model and the Resnet34 architecture with pre-trained weights were used. A learning rate decay strategy based on the "one-cycle" method was used to improve the model's convergence. This is a dynamic learning strategy in which the learning rate varies between an upper and a lower bound. A combined loss function based on Dice and Cross-Entropy (CE) losses was used. Dice is a measure of the overlap between the correctly predicted cracks and the total number of pixels. Therefore, Dice typically performs well on datasets that are heavily unbalanced in terms of the number of pixels belonging to the object of interest, i.e., cracks in this case, and the background. During the evaluation of the Dice score, only pixels that belong to cracks were considered. Note that for a single-class, Dice and F1 score are mathematically identical.  In the model, the Dice loss function was combined with the CE loss, which focuses on the model's performance on a pixel-by-pixel basis. The combined loss function was based on weights for the $ \mathrm{Dice}$ and $\mathrm{CE}$ loss functions, and the sum should be equal to 1, i.e. $w_\mathrm{dice} + w_\mathrm{ce} = 1$. Here, $w_\mathrm{dice}=0.5$ and  $w_\mathrm{ce}=0.5$ were used.

As seen in Table \ref{tab:data_pixels}, the percentage of pixels belonging to a crack is less than 0.5\%. To increase the balance of the dataset, training data was selected to ensure that an equal amount of images with cracks and without cracks were used. Redundant images with no cracks were removed from the training phase. To ensure that the same images were used throughout the training, validation and testing phase, the balanced datasets were stored as .csv files. These are provided in the \textbf{2\_model\_input} folder, which includes a list of "crack" and "no-crack" images for each tunnel, as well as a splitting of the dataset into training, validation and testing using a 70/20/10 split.  Moreover, to handle the class imbalance between cracks and background, a weight factor was introduced to penalise errors on the cracks more than the background. Initial testing of the model was conducted to determine suitable values for the maximum learning rate, number of epochs, and the weight of the crack class.

During training, the best model was evaluated based on the Intersection over Union (IoU). The IoU is defined as the number of True Positive (TP) pixels divided by the sum of TP, False Positive (FP) and False Negative (FN). Thus, IoU neglects accurately classified background pixels and focuses more on the accuracy of crack detection compared to the Dice metric. Other metrics that are tracked are Recall and F1 (Dice) score (Equation \ref{eq:metrics}).  

\begin{equation}
   \mathrm{IoU}= \dfrac{TP}{TP + FP +FN} \\
    \mathrm{Recall} = \dfrac{TP}{TP + FN} \\
    \mathrm{F1} = \dfrac{2 TP}{2TP + FP + FN}
    \label{eq:metrics}
\end{equation}

\begin{figure}[h!]

        \begin{subfigure}[b]{0.45\textwidth}
        \centering
    \includegraphics[width=80mm]{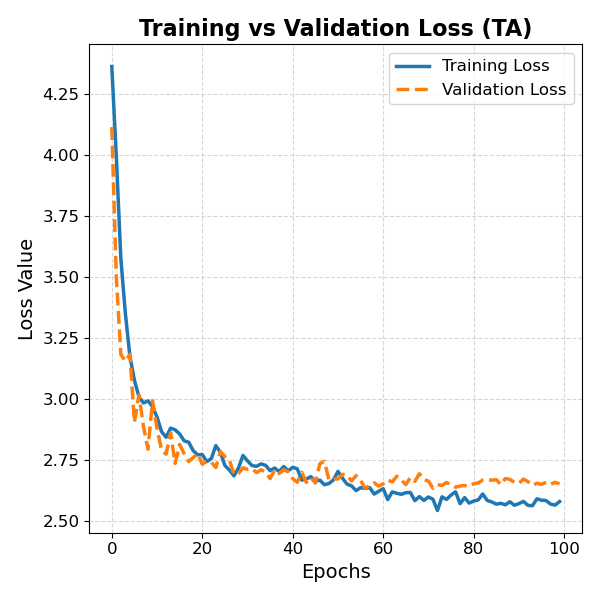}
        \caption{}
        \label{fig:TA_loss}
    \end{subfigure}
    \hfill 
        \begin{subfigure}[b]{0.45\textwidth}
        \centering    
    \includegraphics[width=80mm]{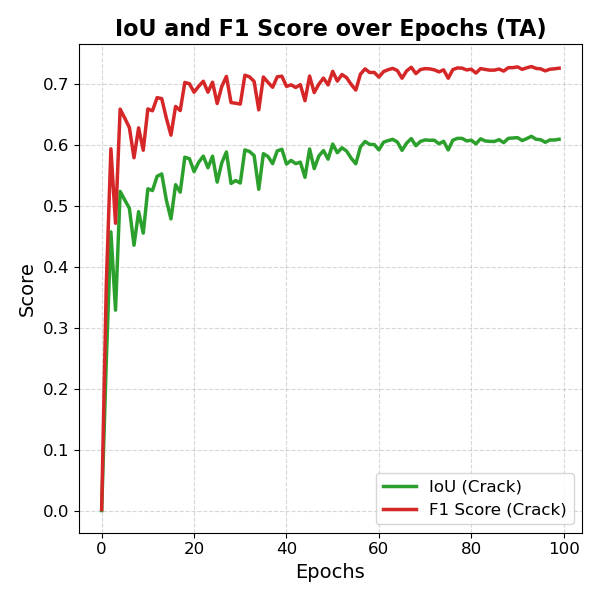}
            \caption{}
        \label{fig:TA_F1}
    \end{subfigure}
	\caption{Visualisation of loss for training and validation during training on Tunnel A (a) and IoU and F1 for Tunnel A (b).}
	\label{fig:training}
\end{figure}

The loss during training and validation for Tunnel A is shown in Figure \ref{fig:training}, together with IoU and F1 score on the validation dataset. It is essential to note that the primary purpose of this model is to provide technical validation of the data, i.e., to verify the quality of the presented dataset and its suitability for training a CNN model. The purpose was not to maximise the performance, i.e. the IoU for the model. The loss, IoU and F1 have all converged, indicating that the number of epochs is sufficient. The value of the loss is high, which is caused by the weight factor used on the crack class, set to 20 in the model. 

\subsection*{Technical challenge}
\begin{figure}[h!]
    \centering
    
    \begin{subfigure}[b]{1\textwidth}
        \centering
        \includegraphics[width=\linewidth]{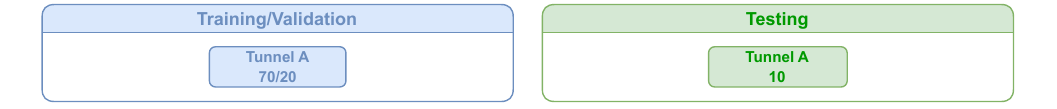}
        \caption{}
        \label{fig:Training_test1}
    \end{subfigure}

    \vspace{1em} 
    
    \begin{subfigure}[b]{1\textwidth}
        \centering
        \includegraphics[width=\linewidth]{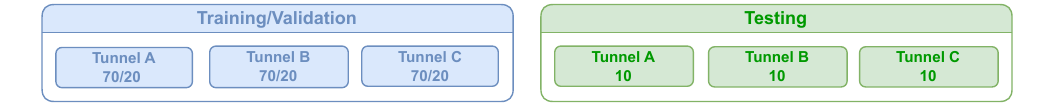}
        \caption{}
        \label{fig:Training_test2}
    \end{subfigure}
    
    \begin{subfigure}[b]{1\textwidth}
        \centering
        \includegraphics[width=\linewidth]{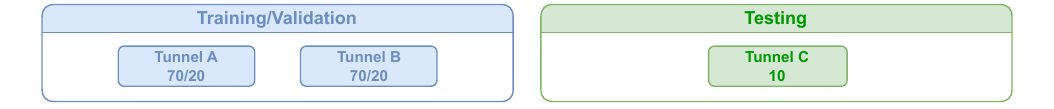}
        \caption{}
        \label{fig:Training_test3}
    \end{subfigure}
    
    \caption{Random split on single domain (a), random split on multi domains (b) and leave-one-out split (domain shift) (c) for training, validation and testing. }
    \label{fig:damage}
\end{figure}

As mentioned, no publicly available dataset with labelled cracks for segmentation tasks in tunnels exists. Papers focusing on crack detection in a tunnel environment have typically collected their own data and created a dataset from one tunnel and thereafter tested the performance of the dataset on the same tunnel. Thus, the model performance in a new domain, i.e. an unseen tunnel, is rarely or never tested. Moreover, the typical small-sized datasets used for training will likely introduce a problem with generalisation of the model, i.e. its ability to perform well on unseen data.  
In this paper, we trained and tested a U-Net model in single and multi-domain scenarios (Figure \ref{fig:Training_test1} and Figure \ref{fig:Training_test2}). Then, to evaluate the ability of the U-Net model to generalize to unseen data, leave-one-out split was used (Figure \ref{fig:Training_test3}), and the following experiments were carried out: 

\begin{itemize}
    \item Training/validation on Tunnel A and B, testing on Tunnel C;
    \item Training/validation on Tunnel A and C, testing on Tunnel B;
    \item Training/validation on Tunnel B and C, testing on Tunnel A.
\end{itemize}

\begin{figure}[h!]
    \centering
    \begin{subfigure}[b]{0.45\textwidth}
        \centering
        \includegraphics[trim={0 0 0 0.7cm},clip, width=1\linewidth]{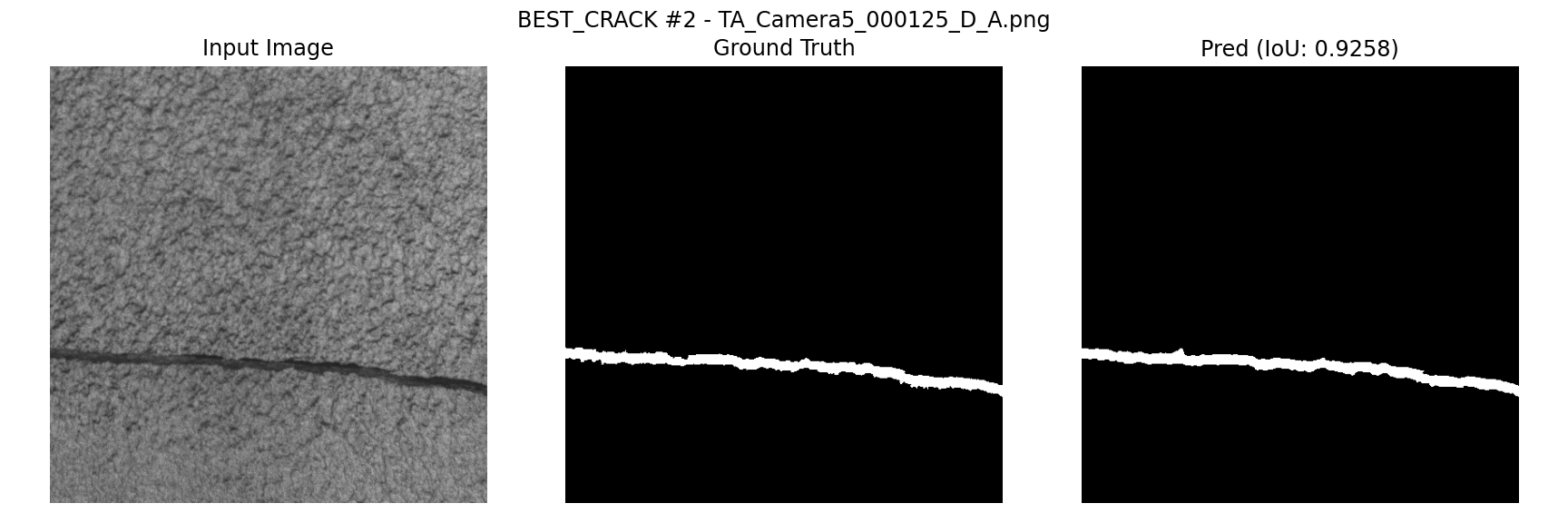}
        \caption{Tunnel A}
        \label{fig:TA_B}
    \end{subfigure}
    \hfill 
    \begin{subfigure}[b]{0.45\textwidth}
        \centering
        \includegraphics[trim={0 0 0 0.7cm},clip, width=1\linewidth]{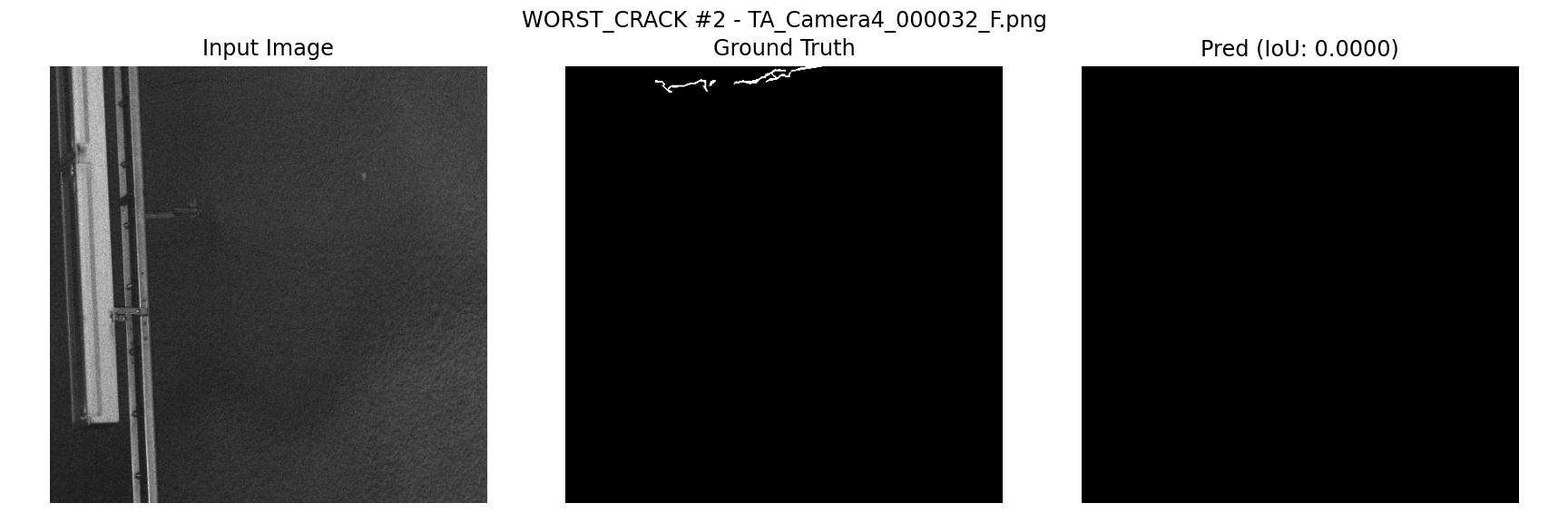}
        \caption{Tunnel A}
        \label{fig:TA_W}
    \end{subfigure}
    
    \begin{subfigure}[b]{0.45\textwidth}
        \centering
        \includegraphics[trim={0 0 0 0.7cm},clip, width=1\linewidth]{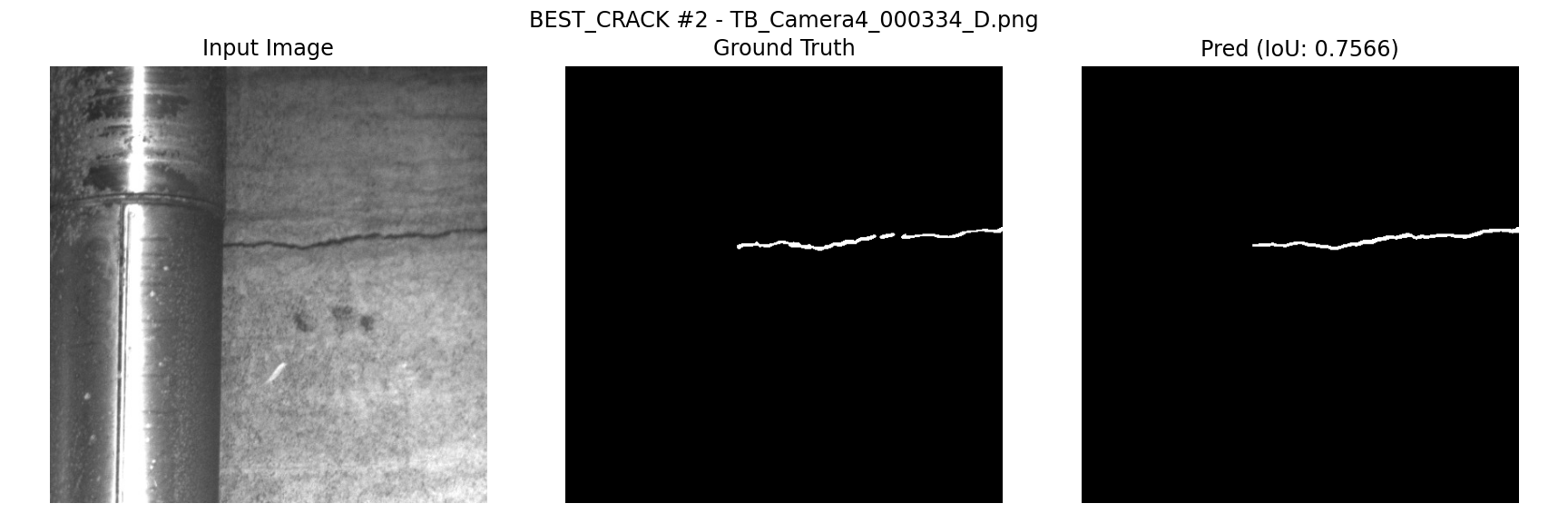}
        \caption{Tunnel B}
        \label{fig:TB_B}
    \end{subfigure}
    \hfill
        \begin{subfigure}[b]{0.45\textwidth}
        \centering
        \includegraphics[trim={0 0 0 0.7cm},clip, width=1\linewidth]{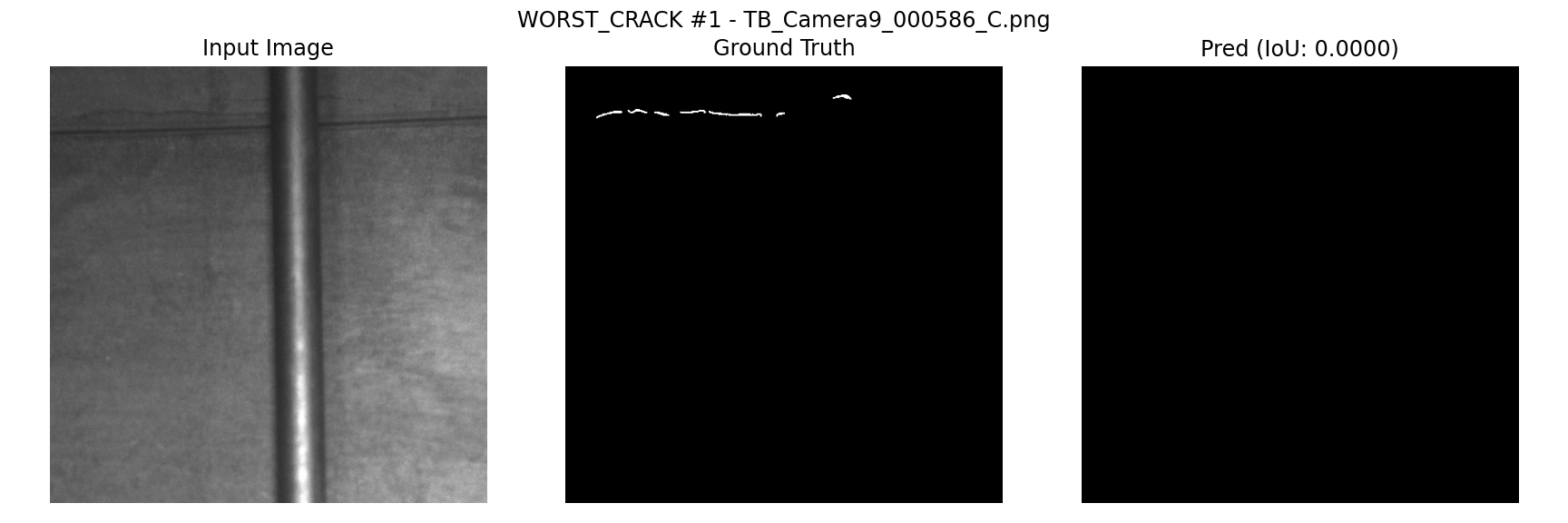}
        \caption{Tunnel B}
        \label{fig:TB_W}
    \end{subfigure}

        \begin{subfigure}[b]{0.45\textwidth}
        \centering
        \includegraphics[trim={0 0 0 0.7cm},clip, width=1\linewidth]{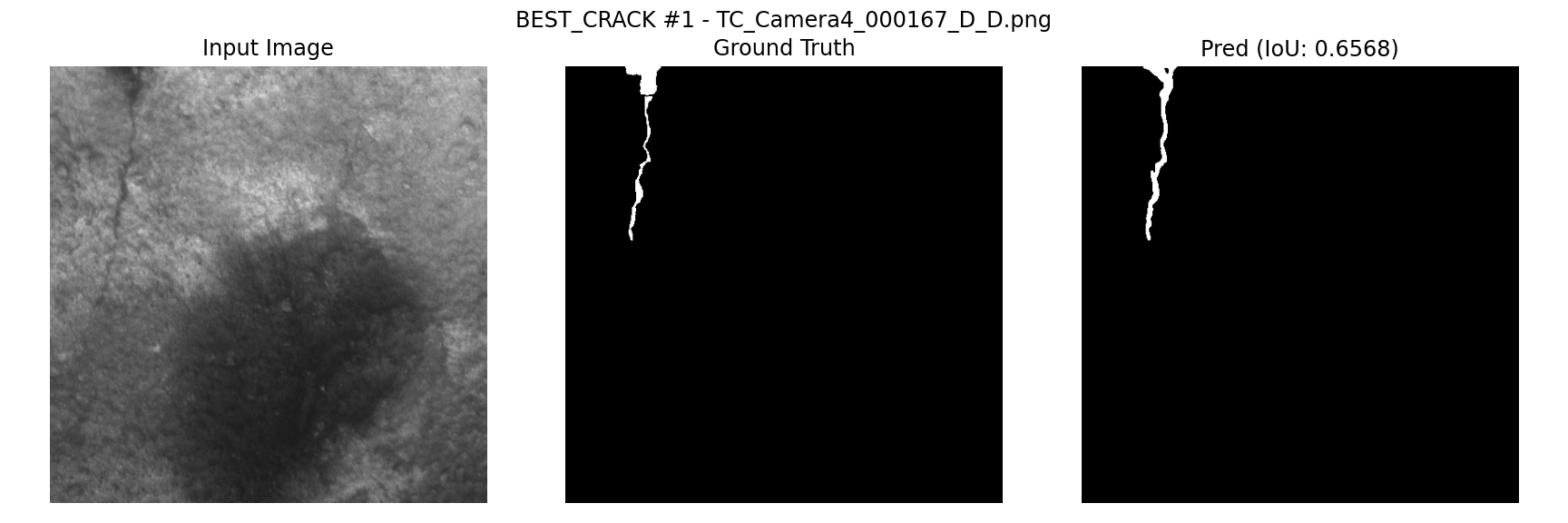}
        \caption{Tunnel C}
        \label{fig:TC_B}
    \end{subfigure}
    \hfill
        \begin{subfigure}[b]{0.45\textwidth}
        \centering
        \includegraphics[trim={0 0 0 0.7cm},clip, width=1\linewidth]{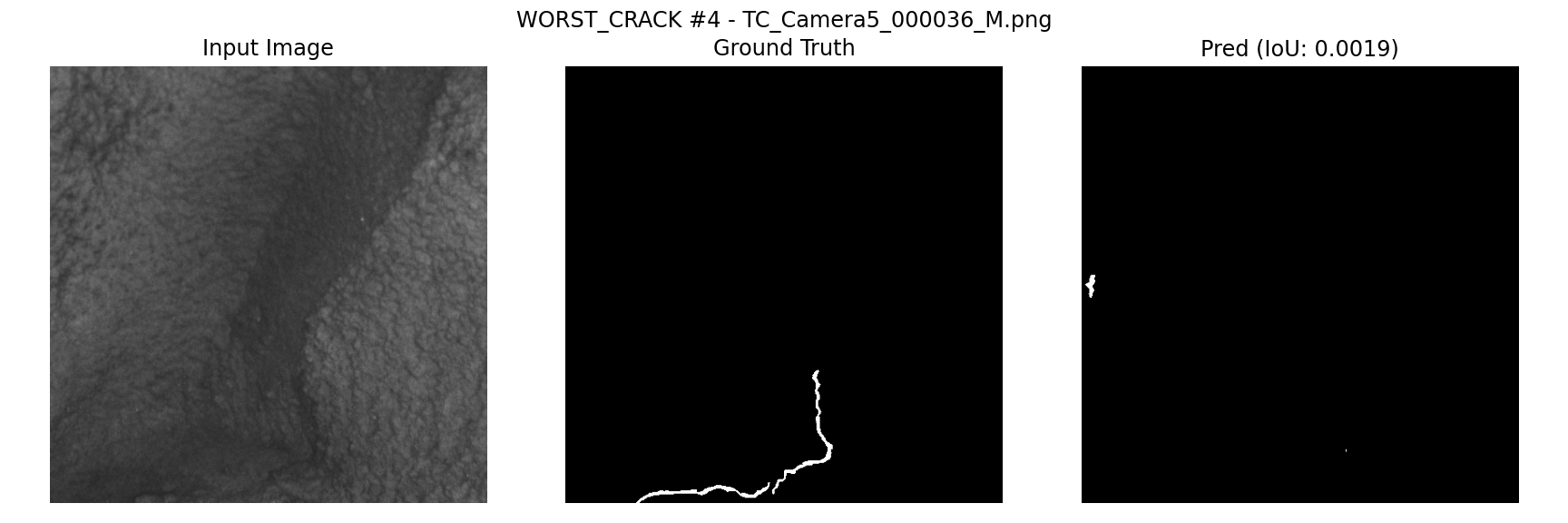}
        \caption{Tunnel C}
        \label{fig:TC_W}
    \end{subfigure}   
    \caption{Best (a, c, e) and worst  (b, d, f) segmentation results for single domain test on Tunnel A, Tunnel B and Tunnel C.}
    \label{fig:11}
\end{figure}

For the single-domain tests, data from the same tunnel were used for training, validation, and testing.
Examples of the best and worst predictions are shown in Figure \ref{fig:11}.

For multi-domain tests, data from all tunnels were used. For the sake of consistency, the same data was used for training, evaluation and testing with a split of 70/20/10. This was ensured by the csv files available in \textbf{2\_model\_input}. Moreover, this also ensured that a balanced dataset with an equal number of images with and without cracks was used. For the domain shift study, training and validation were performed using data from two tunnels, while testing was conducted using data from the unseen tunnel. Here, the same data used for single-domain training were employed, e.g. training and validation data for Tunnels A and B were combined with testing data for Tunnel C. Finally, testing was also performed using the entire dataset of the tunnel that was excluded from the training and validation processes. The metrics are presented in Table \ref{tab:eva_pixel}. 

As shown in Table \ref{tab:eva_pixel}, the metrics during validation and testing are quite similar for both single- and multi-domain models, i.e., when the model is trained and tested using data from the same domains. When a domain shift is introduced, the models' capability of predicting cracks on the new domain, i.e. the test dataset, drops significantly. An exception exists, which is when the model is tested using data from Tunnel A. In this case, the metrics are higher during testing compared to validation. The most likely case is that cracks in Tunnel A are easier to detect. An explanation for this could be the larger size of the cracks, as evident in the histogram plot in Figure \ref{fig:histogram} and the example images shown in Figure \ref{fig:damage_1}. Using part of the data, i.e. 10\%, or the full dataset during testing seems to have a negligible effect since only the metrics for testing Tunnel C are affected. This is likely an indication of a slight imbalance between how easy the cracks are to detect and highlights the importance of running a full k-fold validation during training to investigate this.

Again, the purpose of the model was not to achieve the highest performance but to give a technical proof of concept that the labelled dataset can be used to train a CNN model to detect cracks. Furthermore, the dataset presents an important challenge for tunnel inspection based on DL methods: the transferability of the trained model. The dataset can be used to study how various model architectures and splitting of training data could improve the transferability of the model.


\begin{table}[ht]
\centering
\begin{tabular}{l|lccc|lccc}

\textbf{Domain}  & \textbf{Training/Validation} & \multicolumn{3}{c}{\textbf{Validation}} & \textbf{Testing} & \multicolumn{3}{c}{\textbf{Testing}}  \\
& \textbf{(70/20\%)} &  \textbf{F1} & \textbf{IoU} & \textbf{Recall} & \textbf{} &
\textbf{F1 } & \textbf{IoU} & \textbf{Recall}   \\
\hline
\textbf{Single} & TA & 0.73 & 0.61 & 0.78   & TA (10\%) & 0.85 & 0.75 & 0.84 \\
\textbf{Single} & TB & 0.58 & 0.42 & 0.65  & TB (10\%) & 0.45 & 0.30 & 0.43 \\
\textbf{Single} & TC & 0.43 & 0.29 & 0.47  & TC (10\%) & 0.53 & 0.36 & 0.65  \\ 
\textbf{Multi}& TA + TB + TC & 0.61 & 0.48 & 0.66  & TA + TB + TC (10\%) & 0.65 & 0.52 & 0.70\\ 
\textbf{Shift} & TA + TB & 0.66 & 0.52 & 0.76 & TC (10\%) & 0.54 & 0.38 & 0.64\\
\textbf{Shift} & TA + TB & 0.66 & 0.52 & 0.76 & TC (100\%)  & 0.43 & 0.28 & 0.46\\
\textbf{Shift} & TA + TC & 0.63 & 0.50 & 0.69 & TB (10\%) & 0.40 & 0.26 & 0.48 \\
\textbf{Shift} & TA + TC & 0.63 & 0.50 & 0.69 & TB (100\%) & 0.37 & 0.24 & 0.52 \\
\textbf{Shift}& TB + TC & 0.54 & 0.38 & 0.62 & TA (10\%) & 0.71 & 0.56 & 0.67\\
\textbf{Shift}& TB + TC & 0.54 & 0.38 & 0.62 & TA (100\%) & 0.67 & 0.53 & 0.62\\
\hline
\end{tabular}
\caption{\label{tab:eva_pixel}  F1, IoU and Recall during validation and testing on unseen data.}
\end{table}

\section*{Usage Notes}

The datasets were uploaded to HuggingFace as of December 5, 2025. Any possible modifications, e.g. uploading more data, are declared there.  

\section*{Data Availability}
The TTD dataset is available on Hugging Face via the link \href{https://huggingface.co/datasets/TACK-project/TACK_Tunnel_Data}{https://huggingface.co/datasets/TACK-project/TACK\_Tunnel\_Data}.

\section*{Code Availability}
Image labelling was performed with the SuperAnnotate online platform \cite{SuperAnnotate2024}.
For technical validation, a deep learning model was trained using the \textit{fastai} and \textit{torch} packages. This code was produced with the help of generative AI using Google's Gemini Pro platform. The code to create a classification dataset and to train and evaluate the performance of a CNN model using the U-Net architecture are available in the \textbf{1\_python} folder.  

\section*{Acknowledgements}

This project has received funding from the Richerts Foundation, managed by Sweco Sweden, and Formas - A Swedish research council for sustainable development. The authors acknowledge the SuperAnnotate team, who made the software available for labelling. The authors thank WSP Sweden for collecting the data. Finally, the authors acknowledge the Swedish Transport Administration and Stockholm Region for their willingness to share these unique datasets. 

\section*{Author Contributions Statement}
\textbf{Andreas Sjölander:} Conceptualization, Data Curation, Methodology, Writing - Original Draft, Project administration. \textbf{Valeria Belloni:} Conceptualization, Methodology, Writing - Original Draft. \textbf{Robel Fekadu:} Data Curation. \textbf{Andrea Nascetti:} Conceptualization, Methodology, Writing - Review \& Editing.

\section*{Competing interests} 
The authors declare no competing interests.

\bibliography{reference}

\end{document}